\documentclass[a4paper, 10pt, conference]{ieeeconf}      
\usepackage{FG2021}

\FGfinalcopy 

\IEEEoverridecommandlockouts                              
\usepackage[accsupp]{axessibility} 
\usepackage{subfigure}
\usepackage{times}
\usepackage[pagebackref=true, breaklinks=true, colorlinks, bookmarks=false]{hyperref}
\usepackage{graphicx,multirow}
\usepackage{graphicx}
\usepackage{amsmath}
\usepackage{amssymb}
\usepackage{mathtools}
\usepackage{pdflscape}
\usepackage{amssymb}
\usepackage{pifont}
\newcommand{\cmark}{\ding{51}}%
\newcommand{\xmark}{\ding{55}}%
\usepackage{soul,color}
\usepackage{AbdAlmageed_style}
\usepackage{lipsum}

\DeclareMathOperator*{\argmax}{arg\,max}

\usepackage{xspace}

\makeatletter
\DeclareRobustCommand\onedot{\futurelet\@let@token\@onedot}
\def\@onedot{\ifx\@let@token.\else.\null\fi\xspace}

\def\eg{\emph{e.g}\onedot} 
\def\ie{\emph{i.e}\onedot}

\makeatother

\def\FGPaperID{401} 

\title{\LARGE \bf
Explaining Face Presentation Attack Detection Using Natural Language}

\author{\parbox{16cm}{\centering
    {\large Hengameh Mirzaalian, Mohamed E. Hussein$^{\dagger}$\thanks{$^\dagger$ Mohamed E. Hussein is also affiliated with the Faculty of Engineering, Alexandria University, Alexandria, Egypt}, Leonidas Spinoulas, Jonathan May, and \\Wael Abd-Almageed}\\
    {\normalsize
    University of Southern California, Information Sciences Institute, Marina del Rey, CA, USA
    }}
    \thanks{}
}

\begin{document}
\ifFGfinal
\thispagestyle{empty}
\pagestyle{empty}
\else
\author{Anonymous FG2021 submission\\ Paper ID \FGPaperID \\}
\pagestyle{plain}
\fi
\maketitle
\begin{abstract}
A large number of deep neural network based techniques have been developed to address the challenging problem of face presentation attack detection (PAD). Whereas such techniques' focus has been on improving PAD performance in terms of classification accuracy and robustness against unseen attacks and environmental conditions, there exists little attention on the explainability of PAD predictions. In this paper, we tackle the problem of explaining PAD predictions through natural language. Our approach passes feature representations of a deep layer of the PAD model to a language model to generate text describing the reasoning behind the PAD prediction. Due to the limited amount of annotated data in our study, we apply a light-weight LSTM network as our natural language generation model.
We investigate how the quality of the generated explanations is affected by different loss functions, including the commonly used word-wise cross entropy loss, a sentence discriminative loss, and a sentence semantic loss.
We perform our experiments using face images from a dataset consisting of 1,105 bona-fide and 924 presentation attack samples. Our quantitative and qualitative results show the effectiveness of our model for generating proper PAD explanations through text as well as the power of the sentence-wise losses. To the best of our knowledge, this is the first introduction of a joint biometrics-NLP task. Our dataset can be obtained through our GitHub page\footnote{\url{https://github.com/ISICV/PADISI_USC_Dataset}}.
\end{abstract}

\section{INTRODUCTION}
Face-based biometric authentication systems have been widely used since they provide a high level of security and convenience at a low cost. They eliminate the need to carry identification cards or remember complicated passwords. However, face biometric systems are vulnerable to \textit{presentation attacks} (PA); \ie, the presentation of fake biometric samples in order to impersonate an authorized user or obfuscate the identity of an unauthorized user. Face PA instruments (PAIs) can be printed photos, silicone masks, or unusual makeup used to alter a person's appearance.  To maintain the integrity of biometric systems, a number of methods have been developed for accurate and robust \textit{presentation attack detection} (PAD), possibly using special hardware, such as the usage of multispectral sensing to collect distinctive responses for PAIs compared to \emph{bona-fide} samples \cite{spinoulas2020multispectral}, followed by applying a downstream PAD model using either traditional classifiers \cite{LBP2015,Pereira2013,Patel2016,Boulkenafet2015,Komulainen2013,Komulainen2013,SURF2016} or deep neural networks (DNNs)  \cite{Qin2020,Yu2020,Liu2018,yang2014learn,Gan2017,jaiswal2019ropad,jourabloo2018face,atoum2017face,liu2018learning,perez2019deep,yang2019face,yu2020searching,wang2020deep}.

DNNs have led to impressive performance on different tasks compared to traditional models. However, as opposed to traditional models, which are based on a set of interpretable \emph{hand-crafted} features, mainly encoding local relationships among nearby pixels, learnt representations of DNNs are often criticized for the lack of interpretability due to using stacked convolutions and nonlinear activations~\cite{vilone2020explainable}, among other reasons. Therefore, by using  deeper and more complex networks, one effectively sacrifices interpretability for achieving improved task performance.

The wide deployment of DNN models in real world applications hinges on the user's trust, for which understanding the model's behavior for seen and unseen scenarios is critical. On this front, a number of explainable artificial intelligence (XAI) methods have been introduced~\cite{selvaraju2016gradcam,zhou2015learning,zeiler2013visualizing,Fong2017}. To the best of our knowledge, there exists only a few XAI works in the PAD domain, which are either heatmap-based~\cite{williford2020explainable,Silva_RECPAD_2020} or attention-based~\cite{Chen_2021_WACV} techniques. However, these methods do not reveal how the model processes data and none of them provides any semantic information explaining the final predictions made by the network.

Our goal in this work is to produce descriptive explanations for the decisions made by a face-PAD network through natural language. We use a recurrent neural network (RNN) as a language generation model that consumes PAD representations and  generates natural language descriptions. The PAD representations can be generated using any pre-trained face-PAD model. Further, we incorporate a pre-trained language model to provide word-embeddings for the language generation model rather than training a word-embedding layer from scratch. We investigate the usefulness of applying different losses, including word-wise and sentence-wise losses, in enhancing the quality of the generated descriptions. To the best of our knowledge, this is the first introduction of a joint biometrics--natural-language-processing (NLP) task. 

In the following, we start by providing a brief review of existing face-PAD and XAI techniques in \cref{sec:related_works}. Then, in \cref{sec:problem_formulation}, we analyze different formulations of our proposed XFace-PAD model followed by our word-embedding technique and sentence losses. Details of the dataset, metrics, and results are reported in \cref{sec:evaluation_benchmark,sec:results}. Finally, in \cref{sec:conclusion}, we conclude the paper and discuss directions for future work.

\begin{figure*}[t!]
    \centering
    \includegraphics[width=.99\linewidth]{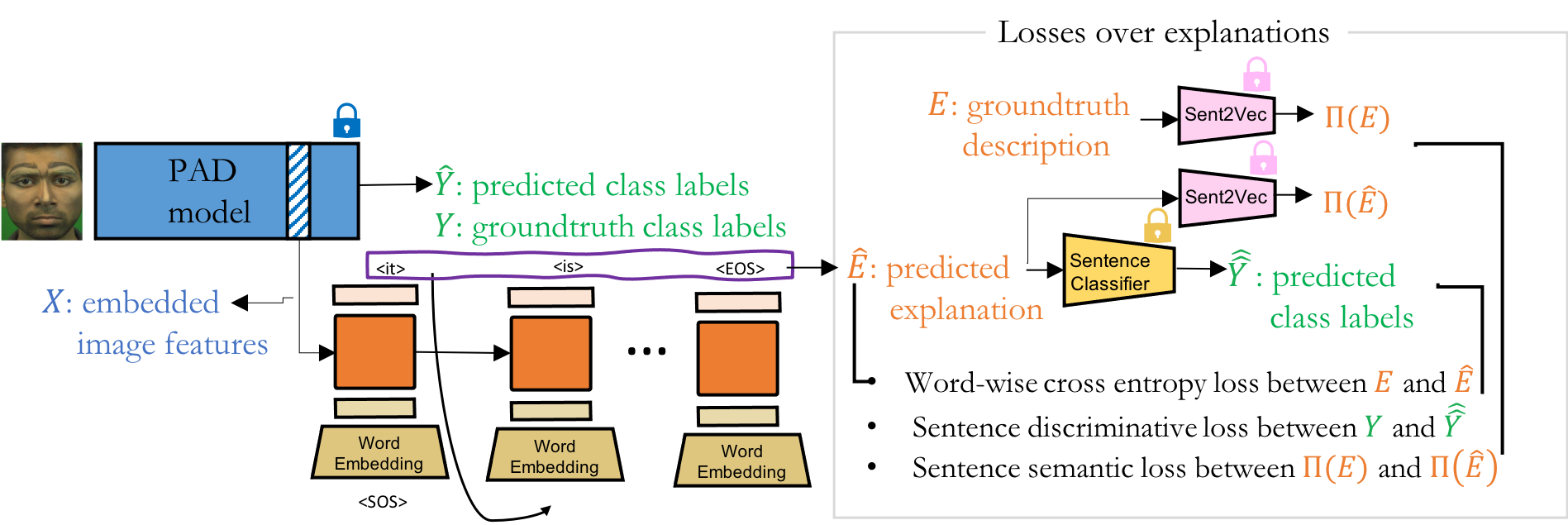}
    \caption{Schematic representation of our proposed explainable face-PAD (XFace-PAD) model through language model. Given a face-PAD model, we pass feature representation of one of its deeper layers to a language model to generate face-PAD descriptions. 
    To maximize the quality of the generated sentences during training the XFace-PAD model, we consider: i) word-wise comparison between the generated and ground truth sentences computed as cross entropy; ii) sentence-wise semantic comparison computed as cosine similarity of the sentence embeddings; iii) sentence discriminative loss (for class-specificity)  (details in \cref{sec:problem_formulation}). Note that in our implementation, while we train the language generation (LG) module, the parameters of the backbone, the PAD module, the sentence classifier, and the model for sentence embedding are frozen. }
         \label{fig:XFPAD}
\end{figure*}

\section{Related Work}
\label{sec:related_works}
\subsection{Face-PAD Detection}
Without loss of generality, existing PAD approaches can be categorized into methods applying traditional classifiers such as support vector machines and random forests  ~\cite{LBP2015,Pereira2013,Patel2016,Boulkenafet2015,Komulainen2013,Komulainen2013,SURF2016} and methods using DNNs~\cite{Qin2020,Yu2020,Liu2018,yang2014learn,Gan2017,jaiswal2019ropad,jourabloo2018face,atoum2017face,liu2018learning,perez2019deep,yang2019face,yu2020searching,wang2020deep}. 

Traditional classifiers rely on extracting a set of \emph{hand-crafted} features mainly encoding local relationships among nearby pixels, such as local binary patterns \cite{LBP2015,Pereira2013}, scale-invariant feature transform \cite{LBP2015,Patel2016}, color texture analysis \cite{Boulkenafet2015,Komulainen2013}, histogram of oriented gradients \cite{Komulainen2013}, and Speeded Up Robust Features \cite{SURF2016}. Deep neural networks, however, consume raw input images and internally extract representations/embeddings learned from the training data. In contrast to traditional classifiers that use interpretable features, embeddings of deeper layers of neural networks are opaque and not easy to interpret, since they are computed out of a number of stacked convolution operations with nonlinear activations.

\subsection{Explainable AI (XAI)}
To address the lack of interpretability of deep neural networks and their black-box nature, explainable artificial intelligence (XAI) methods attempt to generate high-quality interpretable, intuitive, and human-understandable explanations of decisions made by neural networks.

A number of XAI methods are designed with the goal of approximating functions of the CNN networks through local mathematical models such as Taylor expansion~\cite{simonyan2014deep,ribeiro2016why}. 
As opposed to these function-based methods, which are very local and do not measure global interpretable characteristics of the network, heatmap-based methods such as Class Activation Mapping (CAM)~\cite{zhou2015learning} highlight the discriminative parts of the input image detected by the CNN, which contribute more to the final decision made by the network over given query input images. Grad-CAMs~\cite{selvaraju2016gradcam} consitute a generalized version of CAMs, which use the class-specific gradient information flowing into the final convolutional layer of a CNN to produce a coarse localization map of the important regions in the image. Similar to CAM-based techniques~\cite{zhou2015learning,selvaraju2016gradcam}, perturbation based methods~\cite{zeiler2013visualizing,Fong2017} also provide heatmaps, which result from occluding/masking different parts of the image and evaluating the network's performance to discover which part of the input image most affects the predicted output by the network. Rather than spatial attention of the network resulting from heatmap-based methods mentioned above, there exist a number of works on computing channel attentions within the network and ranking the importance of the input channels at each layer of the network through residual channel attentions~\cite{hu2017squeezeandexcitation,chen2016scacnn,zhang2018image}. 

The aforementioned methods have been widely applied in various applications, including biometric data~\cite{Silva_RECPAD_2020,Chen_2021_WACV,williford2020explainable}. However, none of these techniques provide any \emph{semantic information} describing the predictions made by the network. 
One possibility for providing explanations for the predictions made by the network is considering attributes as auxiliary information~\cite{Zhang_2017_CVPR,Lampert_2009_CVPR,Tsai_2017_ICCV,Akata_2013_CVPR}, on which the network jointly predicts class labels and estimates the presence or absence of different attributes in the input image. However, the disadvantage of considering attributes is that they require fine-grained annotations and they need to be revised in the case of having a new class to ensure discrimination among multiple classes. 
To address the lack of generalizability of attribute-based approaches, Hendricks et al.~\cite{hendricks2016generating} employ natural language models to prepare explanations. 

In this paper, we address face PAD explainability (XFace-PAD) using natural language as a biometrics-NLP task. Compared to~\cite{hendricks2016generating}, to avoid overfitting on our relatively small-sized dataset, we use word-embeddings provided by a pre-trained language model rather than training a word-embedding layer from scratch. To maximize the quality of the generated descriptions, we not only include word-wise cross entropy loss, but also consider sentence-wise semantic comparisons computed as cosine similarities over BERT~\cite{devlin-etal-2019-bert} sentence embeddings. The word-wise loss encourages word-alignment between the ground-truth and generated descriptions, whereas the sentence semantic loss encourages the semantic similarity between them. 

\begin{figure}[t!]
    \centering
    \includegraphics[width=\linewidth]{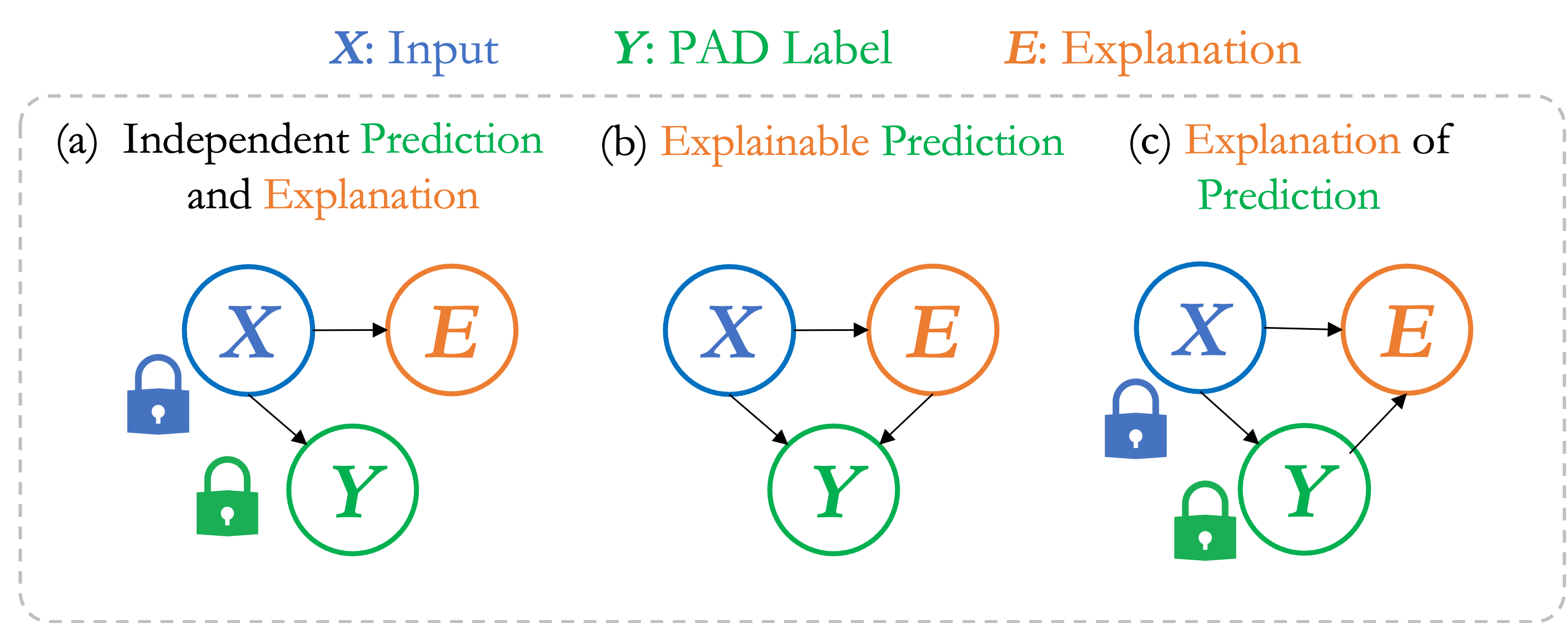}
    \caption{Three different graphical model representations of our pipeline in~\cref{fig:XFPAD}. In (a), the explanation is only conditioned on the feature representation of the input. In (b), the PAD class label is conditioned on both the descriptions as well as the input features. In (c), the description is not only conditioned on the input features but also on the PAD class label.}
         \label{fig:graphical_models}
\end{figure}

\section{Natural Language-Explainable Face PAD}
\label{sec:approach}

\subsection{Problem Formulation}
\label{sec:problem_formulation}
Let us denote the feature representation of an input image by $X$, its corresponding class label by $Y$, and its natural language explanation by $E$. Our objective is to find the predicted class label $\Hat{Y}$ and its generated natural language explanation $\Hat{E}$ that maximize the joint probability distribution $P_{Y,E}$, \ie,
\begin{equation}
\Hat{Y}, \Hat{E} = \argmax_{Y,E} P_{Y,E|X}(Y,E|X). \enspace 
\label{eq:formulation}
\end{equation}
To solve this problem in practice, we need to factorize the probability distribution $P_{Y,E}$. Otherwise, huge amounts of data will be needed to estimate the parameters of the joint distribution. There are different ways for performing this factorization, each of which leads to a conceptually different formulation of the problem. These formulations are depicted as graphical models in~\cref{fig:graphical_models}.

\subsubsection{Independent Prediction and Explanation}
\label{sec:modelA}
The simplest way to factorize the joint probability distribution in \cref{eq:formulation} is to assume  conditional independence between $Y$ and $E$ given $X$. In this case,
\begin{equation}
P^A_{Y,E|X}(Y,E|X) = P_{Y|X}(Y|X)P_{E|X}(E|X) \enspace ,
\label{eq:formulationA}
\end{equation}
which is shown in \cref{fig:graphical_models}(a). In this scenario, if the feature representation $X$ is obtained independently from $Y$ and $E$, \eg, using a pre-trained model, then the PAD predictions and explanations are practically independent. We argue that this model does not capture the essence of what we want to achieve, which is explaining the PAD prediction. In fact, if we only consider the explanation variable of this model, the model will be reminiscent of an \textit{image captioning} model~\cite{he2020image, pmlr-v37-xuc15} in which textual captions are directly generated from the visual input. Nevertheless, this model will serve as a baseline to our work. For simplicity, we will refer to it as model \textit{A} for the remainder of the paper.  

\subsubsection{Explainable Prediction}
\label{sec:modelB} 
Another way to factorize the joint probability distribution in \cref{eq:formulation}
is to add to the model in \cref{sec:modelA} the dependence of the PAD label $Y$ on the explanation $E$, \ie,
\begin{equation}
P^B_{Y,E|X}(Y,E|X) = P_{E|X}(E|X)P_{Y|E,X}(Y|E,X) \enspace .
\label{eq:formulationB}
\end{equation}
This scenario, which is illustrated in \cref{fig:graphical_models}(b) and referred to as Model \textit{B}, captures the situation in which an explanation for the feature representation can assist the PAD prediction. We can also view this as an attempt to making the PAD label prediction explainable by making it dependent on the representation's explanation. This is why we call this case, \textit{explainable prediction}. 
We will not explicitly study this model in this paper and studying it is a part of our future work, as discussed in the conclusion section. 

\subsubsection{Explanation of Prediction}
\label{sec:modelC}
A third way to factorize the joint probability distribution in \cref{eq:formulation}
is to reverse what we did in \cref{sec:modelB} and assume that the explanation is dependent on the class label, as illustrated in \cref{fig:graphical_models}(c), and can be mathematically represented as
\begin{equation}
P^C_{Y,E|X}(Y,E|X) = P_{Y|X}(Y|X)P_{E|Y,X}(E|Y,X) \enspace .
\label{eq:formulationC}
\end{equation}
This model captures our notion about explaining the prediction. The explanation depends on both the feature representation and the PAD label; \ie, the explanation can be considered as an explanation for why the predicted label is the best for the input sample. We refer to this formulation as model \textit{C}.

\subsection{Model Realizations}
Our models are realized via deep neural networks. For an input image $I$, its feature representation $X$ is obtained through a convolutional neural network (CNN). The PAD prediction $\Hat{Y}$ is obtained through a multi-layer perceptron (MLP) parametrized by $\Theta_p$. The explanation is generated using a language generation recurrent neural network (RNN) parametrized by $\Theta_e$. \cref{fig:XFPAD} shows an overview of our entire model including the PAD and language generation (LG) modules.

The distinction between the different graphical model realizations (described in \cref{sec:modelA,sec:modelC}) is achieved through varying the connections between the PAD module and the language generation (LG) module. Let us denote the computed losses at the PAD and LG modules by $\mathcal{L}_{\text{pad}}$ and $\mathcal{L}_{\text{lg}}$, respectively. The two losses are formulated as cross entropy losses, as follows:
\begin{align}
    \mathcal{L}_{\text{pad}}& = - \sum\limits_{n=0}^{N}\log(P_{p}(\Hat{Y_n}=Y_n \vert \mathcal{G}_n; \Theta_p)) \label{eq:pad_loss} \\
     \mathcal{L}_{\text{lg}}^a& = - \sum\limits_{n=0}^{N}\sum\limits_{t=0}^{T}\log(P_{e}(\Hat{E}^t_n=E^t_n \vert \mathcal{H}_n, E^0_n..E^{t-1}_n; \Theta_e))\enspace, \label{eq:lg_loss}
\end{align}
where $N$ is the batch size,  $E^{i}_n$  and $\Hat{E}^{i}_n$ are the $i^{\text{th}}$ word of the ground-truth and predicted sentences, respectively; and $P_{p}$ and $P_e$ are the estimated categorical probability distributions out of the PAD and the LG modules, respectively, from which our loss function directly uses the estimated probabilities of the predicted PAD label $\hat{Y}_n$ and the predicted words $\hat{E}^t_n, \forall 0\leq t \leq T$. $\mathcal{G}_n$ and $\mathcal{H}_n$ are the extra dependencies (\ie, the network inputs) for each of the two modules. For model \textit{A}, $\mathcal{G}_n=\mathcal{H}_n=X_n$, while for model \textit{C}, $\mathcal{G}_n=X_n$ and $\mathcal{H}_n=(X_n, \Hat{Y}_n)$.
Given $\mathcal{L}_{\text{pad}}$ and $\mathcal{L}_{\text{lg}}$, the total loss $\mathcal{L}$ of each of the graphical models can be set as their weighted sum: $\mathcal{L}= \omega_1 \mathcal{L}_{\text{pad}} + \omega_2  \mathcal{L}_{\text{lg}}$.

\subsection{Word-Embedding}
\label{sec:word_embedding}
Every language model includes an embedding layer to map predicted or ground-truth words to numerical vector representations. 
In our work, we applied a lightweight linear word embedding layer of size 128.
Note that applying other pre-trained language models, such as BERT~\cite{devlin-etal-2019-bert}, can be considered as an alternate choice for word-embedding. 
However, we opted to use a simpler model considering that word-embedding, unlike sentence embedding (\cref{sec:losses}), is invoked on every iteration of the language generation model. 


\subsection{Sentence-Wise Embedding and Losses} 
\label{sec:losses}
\begin{figure*}[t!]
    \centering
    \subfigure[Mean of word-embeddings by an embedding layer trained from scratch]{\includegraphics[width=.325\linewidth]{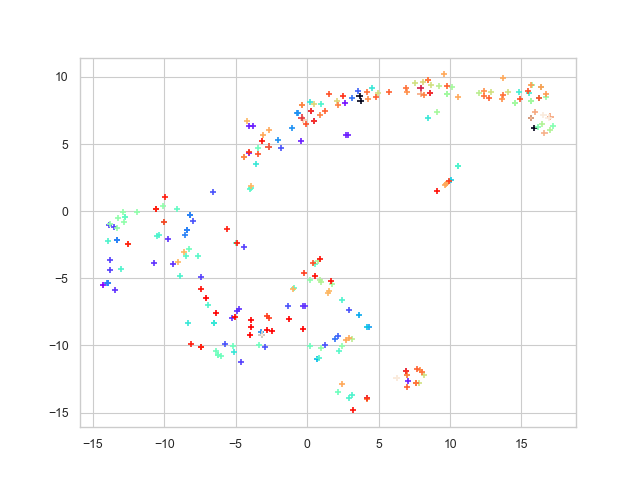}}
    \subfigure[Mean of word-embeddings by Word2Vec \cite{church_2017} ]{\includegraphics[width=.325\linewidth]{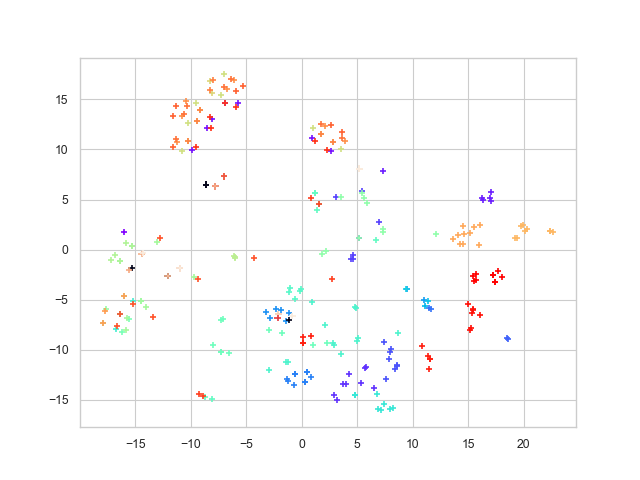}}
    \subfigure[Sentence embeddings by BERT \cite{devlin-etal-2019-bert}.]{\includegraphics[width=.325\linewidth]{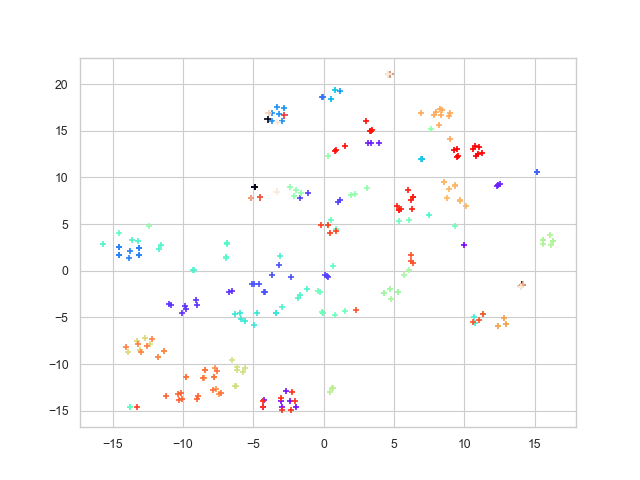}}
    \caption{tSNE visualization of the sentence embeddings by: i) averaging word-embeddings provided by a word-embedding layer of size five trained from scratch over our dataset; ii) averaging  word-embeddings provided by Word2Vec \cite{church_2017} embeddings; and iii) sentence-embeddings provided by BERT \cite{devlin-etal-2019-bert}. Different colors correspond to the different PA types in our dataset. It can be seen that the embeddings provided by pre-trained models such as Word2Vec and BERT are more distinctive among PA types compared to the ones provided by a light weight embedding layer trained from scratch. }
         \label{fig:tsne}
\end{figure*}

Word-wise cross entropy loss is commonly used in training image captioning methods, \eg, \cite{pmlr-v37-xuc15,Rajarshi2020}, which is what we also use in \cref{eq:lg_loss}. We will refer to this loss later as $\mathcal{L}_{ww}$ to distinguish it from other language generation losses to be introduced in this section.
This loss encourages word-wise alignment between generated and ground-truth sentences. However, it ignores contextual information within the sentences \cite{devlin-etal-2019-bert}. Consider, for example, the following pairs of ground truth and corresponding generated explanations:
\begin{itemize}
    \item Pair\#1: \{\color{red} it \color{teal}is \color{blue} bona \color{orange} fide\color{black},   \color{red} bona \color{teal} fide \color{blue} it \color{orange} is\color{black}\}
    \item Pair\#2: \{\color{red} it \color{teal}is \color{blue} bona \color{orange} fide\color{black},  \color{red} plastic \color{teal}mask \color{blue} is \color{orange} presented\color{black}\}
\end{itemize}
The word-wise cross entropy losses for the two pairs can be similar because words in the same sequential order (represented with same color) are compared to each other, whereas it makes more sense to have higher loss for Pair\#2 compared to Pair\#1. To address this issue, we introduce a sentence semantic loss which is based on the cosine similarity between the sentence embeddings. Let us denote sentence embeddings of the ground-truth $E_n$ and generated $\hat{E}_n$ sentences
by $\Pi(E_n)$ and $\Pi(\Hat{E}_n)$, respectively, where $\Pi$ denotes the sentence to vector mapping. Then, we set our sentence semantic loss as:
\begin{equation}
 L_{ss} = - \sum\limits_{n=0}^{N}  \cos(\Pi(E_n), \Pi(\Hat{E}_n))
\label{eq:cos_loss}
\end{equation}
Sentence embedding $\Pi$ in \cref{eq:cos_loss} can be obtained either by averaging word-embeddings of the sentence words or directly through a pre-trained contextualized language model, such as BERT \cite{devlin-etal-2019-bert}.
In \cref{fig:tsne}, we show tSNE visualizations of the sentence embeddings of PA descriptions in our dataset computed by:  (i) averaging word-embeddings provided by a word-embedding layer of size five trained from scratch over our dataset; (ii) averaging  word-embeddings provided by Word2Vec \cite{church_2017} embeddings; and (iii) sentence-embeddings provided by BERT \cite{devlin-etal-2019-bert}.
It can be seen that the embeddings provided by pre-trained models such as Word2Vec and BERT are more distinctive among PAs compared to the ones provided by the embedding layer trained from scratch. In our preliminary experiments, using sentence embeddings based on BERT performed significantly better than sentence embedding using Word2Vec. Therefore, in the results section, whenever the sentence semantic loss $\mathcal{L}_{ss}$ is applied, the BERT model is used to construct the sentence embeddings.

In addition to the sentence semantic loss, we also observed improvement when applying the sentence discriminative loss suggested in~\cite{hendricks2016generating}, which  encourages the network to generate descriptions that are PA-discriminative. To do so, a separate light weight classifier is trained off-line over the ground-truth descriptions of the PAs applying the following cross entropy loss:
\begin{equation}
 L_{disc} = - \sum\limits_{n=0}^{N} \log(P_r(\Hat{\Hat{Y_n}}=Y_n\vert E_n; \Theta_\text{disc})) 
\label{eq:disc_loss}
\end{equation}
where $P_r$ is the probability distribution over PAD labels $Y_n$ given the ground-truth descriptions $E_n$. 
The sentence dicriminativeness model is frozen during the training of the entire XFace-PAD model. However, the gradient of its loss (\cref{eq:disc_loss}) is still back-propagated through the model during training to reward the language generation module when the model's predicted PA class $\Hat{\Hat{Y}}_n$ given the generated description matches the ground truth label $Y_n$, \ie, $\Hat{\Hat{Y}}_n = Y_n$.

\section{Experimental Setup}
\label{sec:evaluation_benchmark}
\subsection{Dataset and Partitioning}
\label{sec:dataset}
We perform our evaluations on RGB images of a new dataset~\cite{Rostami_2021_ICCV} consisting of 1,105 bona-fide samples and 924 PA samples (140 funny glasses, 56 printed paper, 71 mannequin, 27 opaque mask, 232 plastic mask, 206 makeup, 33 silicone mask, 66 paper glasses, and 93 tattoo). Each PA image is annotated five times with natural language sentences describing the sample. Each sentence was generated by a different annotator.

We employ a 3-Fold partitioning to alleviate the bias resulting from a fixed division of the dataset into training, testing, and validation sets. The dataset is divided into three roughly equal sets of samples such that the data of each subject only appears in one set. Then, we create the 3-Fold partitioning by using two sets for training and one set for testing each time. $15\%$ of the training data is separated to create a validation set. The partitioning is done such that the data of each participant appears only either in the training set or the validation set in each fold.

\subsection{Implementation Details}
We use two different DNN models as our feature extraction backbones, namely the MOCO-v1 model~\cite{MOCO_2020_CVPR} and the vision-transformer (ViT) model~\cite{dosovitskiy2020vit}. These two models are pre-trained on different subsets of the ImageNet dataset \cite{deng2009imagenet,russakovskyImageNetLargeScale2015}. MOCO-v1 is based on the ResNet-50 architecture~\cite{he2016resnet}. We used the same variant of ViT used by George and Marcel~\cite{Anjith2021IJCB}. For the PAD module, we add an additional fully connected layer after the last layer of each of them to provide predictions over ten  classes in our case: one class corresponding to bona-fide and nine classes corresponding to the nine available PA types in our dataset. 

In our implementation, we fine-tune the last two fully connected layers of the above mentioned backbone models over our dataset only with the PAD loss. We then freeze the parameters of the PAD module while training the language generation module.  Note that we could choose to fine-tune the PAD module at the same time we train the language generation module. However, this would require extra effort for hyper-parameter tuning to maintain the PAD performance uncompromised by training the language module. 

After performing grid search, we settled on the following hyper-parameter settings for our pipeline: two LSTM layers for the LG module with a hidden state of size 100, feature representation of size 128, 0.5 dropout whenever applicable, and 0.0002 learning rate. Weights of each of the three different losses of the LG module $\{\mathcal{L}_{ww}, \mathcal{L}_{disc}, \mathcal{L}_{ss}\}$, were varied depending on the combination of the different losses in our experiments. We optimize our network using Adam optimizer \cite{Adam} and lr-scheduler with 0.5 decay factor every 20 epochs.

\subsection{Evaluation Metrics}
\label{sec:metrics}
Performance of the PAD module is reported in terms of area under the Receiver Operating Characteristic (ROC) curve, $AUC$, and Equal Error Rate (EER). Evaluations of generated descriptions by the language generation module are computed in terms of BLEU \cite{bleu_metric}, METEOR \cite{meteor_metric},  and ROUGE \cite{lin-2004-rouge} metrics. Note that, METEOR is computed by matching words using WordNet \cite{Miller1990}, which includes matching among synonyms.

\section{Results} 
\label{sec:results}
\begin{figure}[t]
    \centering
    \includegraphics[width=\linewidth]{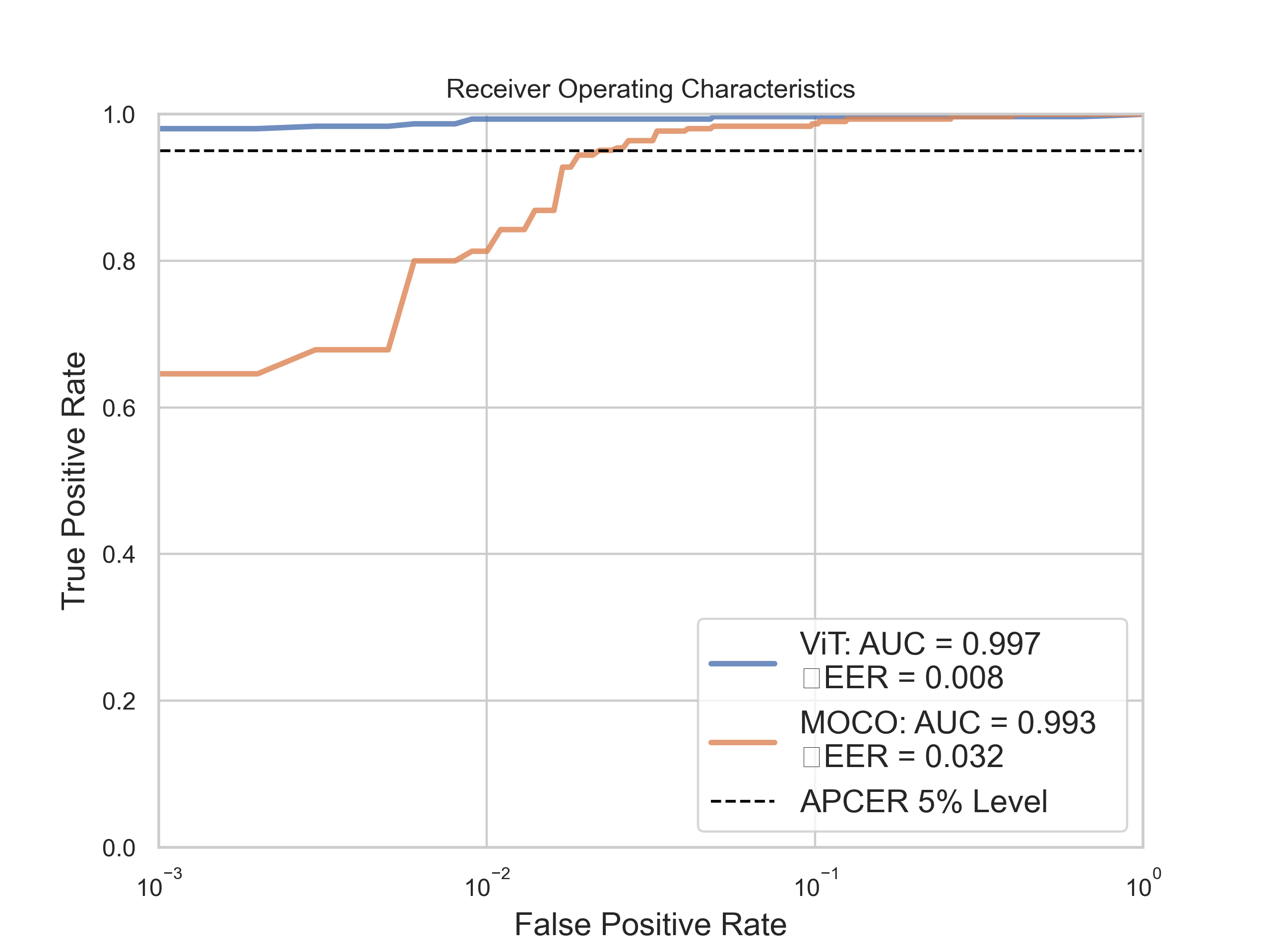}
    \caption{The PAD performance using the MOCO model \cite{MOCO_2020_CVPR} and the ViT model \cite{dosovitskiy2020vit}.}
         \label{fig:ROC}
\end{figure}

\cref{fig:ROC} shows the performance of the PAD module: AUC of 0.993 and 0.997 for the MOCO model \cite{MOCO_2020_CVPR} and the ViT model \cite{dosovitskiy2020vit}, respectively. In our implementation, PAD performance is the same in the presence or absence of the LG module since parameters of the PAD module are frozen while tuning the LG module. 

\begin{figure*}[t!]
    \centering
    \includegraphics[width=.99\linewidth]{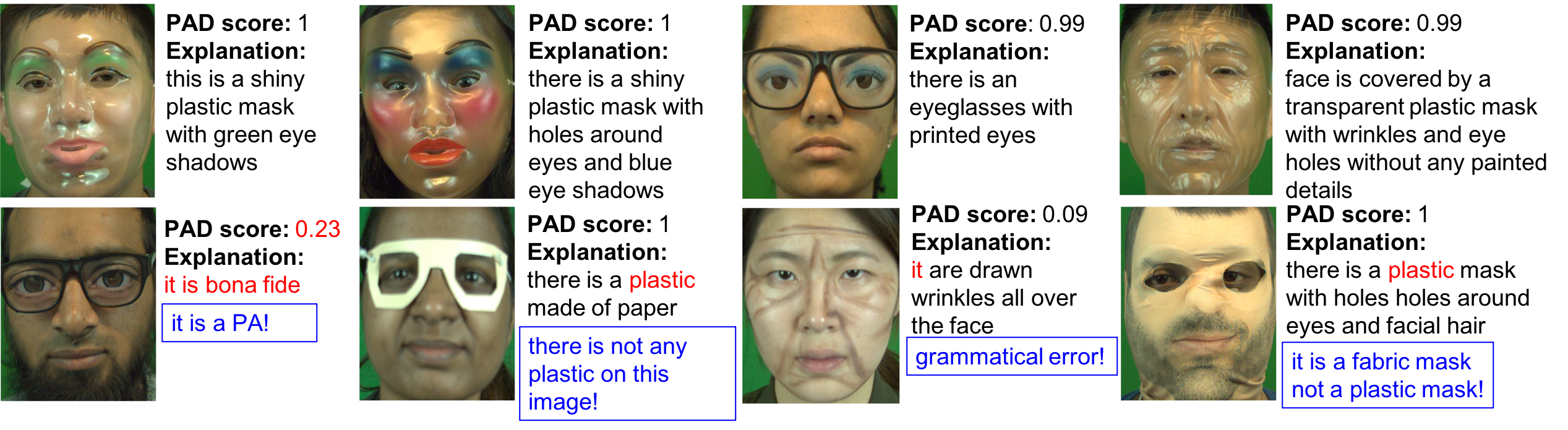}
    \caption{Examples of the generated descriptions over a number of PA samples by our XFace-PAD model. The first row shows meaningful generated descriptions and the second row shows examples of generated descriptions with errors depicted by red text. The blue text provides a comment on the erroneous parts of the generated descriptions. Note that the first image of the second row has a low predicted PA probability, which can justify having the wrong description. }
         \label{fig:ex_desc}
\end{figure*}

\begin{table*}
\caption{Evaluation of the generated descriptions of PA images in terms of BLEU, METEOR, and ROUGE metrics. Since parameters of the PAD model are frozen while training the language generation model, the PAD performance of all the settings is the same (see \cref{fig:ROC}). Each row of the sub-tables represents the results given the graphical model of type \textit{A} (the first row) or type \textit{C} (the second-to-sixth rows) in the presence or absence of the sentence-discriminative loss $\mathcal{L}_{disc}$ or sentence semantic loss $\mathcal{L}_{ss}$. Average of the numerical results over all PAs is reported on the first sub-table (All PAs) and numerical results per PA type (funny glasses, printed paper, mannequin, opaque mask, plastic mask, makeup, silicone mask, paper glasses, and tattoo) are represented in the remaining sub-tables.}
\label{tab:results}
\centering
\begin{tabular}{p{0.3cm}p{2cm} p{0.5cm}p{0.5cm}p{0.5cm} p{1.4cm}p{1.4cm}p{1.4cm}p{1.4cm}p{1.4cm}p{1.4cm}p{1.4cm}p{1.4cm}|}\hline
& Model  & GM &$L_{disc}$ & $L_{ss}$  & BLEU1 & BLEU2 & BLEU3 & METEOR &  ROUGE   \\    \hline
\multirow{6}{*}{\rotatebox[origin=c]{90}{{All PAs}}} 
& MOCO  \cite{MOCO_2020_CVPR} &   a   & \xmark    & \xmark   & 0.46$\pm$0.35 & 0.35$\pm$0.38 & 0.28$\pm$0.39 & 0.44$\pm$0.33 & 0.44$\pm$0.33 \\   
& MOCO  \cite{MOCO_2020_CVPR} &   c  & \xmark    & \xmark     & 0.73$\pm$0.35 & 0.66$\pm$0.40 & 0.62$\pm$0.42 & 0.69$\pm$0.35 & 0.70$\pm$0.34 \\    
& MOCO  \cite{MOCO_2020_CVPR} &   c  & \cmark    & \xmark     & 0.75$\pm$0.34 & 0.69$\pm$0.39 & 0.64$\pm$0.42 & 0.72$\pm$0.34 & 0.72$\pm$0.33 \\    
& MOCO  \cite{MOCO_2020_CVPR} &   c  & \xmark    & \cmark     & 0.79$\pm$0.32 & 0.74$\pm$0.37 & 0.70$\pm$0.40 & 0.76$\pm$0.32 & 0.76$\pm$0.32 \\    
& MOCO  \cite{MOCO_2020_CVPR} &   c  & \cmark    & \cmark     &  {0.80$\pm$0.31} & {0.75$\pm$0.35} & {0.70$\pm$0.39} & {0.77$\pm$0.31} & {0.77$\pm$0.31} \\   
& ViT \cite{dosovitskiy2020vit} & c & \cmark    & \cmark & 0.84$\pm$0.27 & 0.79$\pm$0.31 & 0.75$\pm$0.35 & 0.79$\pm$0.27 & 0.81$\pm$0.27 \\   \hline
\end{tabular}
\vspace{1mm} \\
\begin{tabular}{p{0.3cm}p{2cm} p{0.5cm}p{0.5cm}p{0.5cm} p{1.4cm}p{1.4cm}p{1.4cm}p{1.4cm}p{1.4cm}p{1.4cm}p{1.4cm}p{1.4cm}|}\hline
\multirow{6}{*}{\rotatebox[origin=c]{90}{Funny Glasses}} 
& MOCO  \cite{MOCO_2020_CVPR} &   a  & \xmark    & \xmark     &  0.20$\pm$0.13 & 0.03$\pm$0.07 & 0.00$\pm$0.00 & 0.18$\pm$0.07 & 0.23$\pm$0.11   \\    
& MOCO  \cite{MOCO_2020_CVPR} &   c  & \xmark    & \xmark     &  0.52$\pm$0.31 & 0.41$\pm$0.32 & 0.32$\pm$0.33 & 0.46$\pm$0.27 & 0.51$\pm$0.24\\   
& MOCO  \cite{MOCO_2020_CVPR} &   c  & \cmark    & \xmark     &  0.51$\pm$0.30 & 0.38$\pm$0.31 & 0.29$\pm$0.32 & 0.45$\pm$0.26 & 0.51$\pm$0.23 \\   
& MOCO  \cite{MOCO_2020_CVPR} &   c  & \xmark    & \cmark     & 0.59$\pm$0.33 & 0.49$\pm$0.38 & 0.42$\pm$0.39 & 0.55$\pm$0.32 & 0.58$\pm$0.29\\    
& MOCO  \cite{MOCO_2020_CVPR} &   c  & \cmark    & \cmark     & {0.61$\pm$0.30 }& {0.51$\pm$0.34} & {0.44$\pm$0.36} & {0.55$\pm$0.30} & {0.59$\pm$0.27} \\  
& ViT \cite{dosovitskiy2020vit} & c & \cmark    & \cmark & 0.66$\pm$0.30 & 0.60$\pm$0.30 & 0.54$\pm$0.30 & 0.63$\pm$0.26 & 0.68$\pm$0.24  \\   \hline
\end{tabular}
\vspace{1mm} \\
\begin{tabular}{p{0.3cm}p{2cm} p{0.5cm}p{0.5cm}p{0.5cm} p{1.4cm}p{1.4cm}p{1.4cm}p{1.4cm}p{1.4cm}p{1.4cm}p{1.4cm}p{1.4cm}|}\hline
\multirow{6}{*}{\rotatebox[origin=c]{90}{Printed Paper}} 
& MOCO  \cite{MOCO_2020_CVPR} &   a  & \xmark    & \xmark     &  {1.00$\pm$0.00} & {1.00$\pm$0.00 }& {1.00$\pm$0.00} & {1.00$\pm$0.00} & {1.00$\pm$0.00} \\   
& MOCO  \cite{MOCO_2020_CVPR} &   c  & \xmark    & \xmark     &  0.97$\pm$0.10 & 0.96$\pm$0.14 & 0.93$\pm$0.23 & 0.97$\pm$0.08 & 0.97$\pm$0.11 \\   
& MOCO  \cite{MOCO_2020_CVPR} &   c  & \cmark    & \xmark     & 0.97$\pm$0.10 & 0.96$\pm$0.14 & 0.93$\pm$0.23 & 0.97$\pm$0.08 & 0.97$\pm$0.11 \\   
& MOCO  \cite{MOCO_2020_CVPR} &   c  & \xmark    & \cmark     & {1.00$\pm$0.00} & {1.00$\pm$0.00} & {1.00$\pm$0.00} & {1.00$\pm$0.00} & {1.00$\pm$0.00} \\   
& MOCO  \cite{MOCO_2020_CVPR} &   c  & \cmark    & \cmark     &  {1.00$\pm$0.00} & {1.00$\pm$0.00} & {1.00$\pm$0.00 }& {1.00$\pm$0.00} & 1.00$\pm$0.00 \\   
& ViT \cite{dosovitskiy2020vit} & c & \cmark    & \cmark &  1.00$\pm$0.00 & 1.00$\pm$0.00 & 1.00$\pm$0.00 & 1.00$\pm$0.00 & 1.00$\pm$0.00 \\   \hline
\end{tabular}
\vspace{1mm} \\
\begin{tabular}{p{0.3cm}p{2cm} p{0.5cm}p{0.5cm}p{0.5cm} p{1.4cm}p{1.4cm}p{1.4cm}p{1.4cm}p{1.4cm}p{1.4cm}p{1.4cm}p{1.4cm}|}\hline
\multirow{6}{*}{\rotatebox[origin=c]{90}{Mannequin}} 
& MOCO  \cite{MOCO_2020_CVPR} &   a  & \xmark    & \xmark     & 0.90$\pm$0.09 & 0.84$\pm$0.14 & 0.75$\pm$0.23 & 0.90$\pm$0.09 & 0.89$\pm$0.11  \\    
& MOCO  \cite{MOCO_2020_CVPR} &   c  & \xmark    & \xmark     & 0.88$\pm$0.25 & 0.85$\pm$0.29 & 0.81$\pm$0.31 & 0.87$\pm$0.24 & 0.86$\pm$0.26 \\   
& MOCO  \cite{MOCO_2020_CVPR} &   c  & \cmark    & \xmark     & 0.88$\pm$0.25 & 0.85$\pm$0.29 & 0.81$\pm$0.31 & 0.87$\pm$0.24 & 0.86$\pm$0.26 \\   
& MOCO  \cite{MOCO_2020_CVPR} &   c  & \xmark    & \cmark     & {0.97$\pm$0.07} & {0.95$\pm$0.10} & 0.92$\pm$0.15 & {0.96$\pm$0.08} & {0.95$\pm$0.10}\\    
& MOCO  \cite{MOCO_2020_CVPR} &   c  & \cmark    & \cmark     & {0.97$\pm$0.07} & {0.95$\pm$0.11} & {0.93$\pm$0.17 }& {0.96$\pm$0.10} & {0.95$\pm$0.11} \\  
& ViT \cite{dosovitskiy2020vit} & c & \cmark    & \cmark & 0.97$\pm$0.08 & 0.95$\pm$0.10 & 0.92$\pm$0.15 & 0.94$\pm$0.10 & 0.94$\pm$0.11\\   \hline
\end{tabular}
\vspace{1mm} \\
\begin{tabular}{p{0.3cm}p{2cm} p{0.5cm}p{0.5cm}p{0.5cm} p{1.4cm}p{1.4cm}p{1.4cm}p{1.4cm}p{1.4cm}p{1.4cm}p{1.4cm}p{1.4cm}|}\hline
\multirow{6}{*}{\rotatebox[origin=c]{90}{Opaque Mask}} 
& MOCO  \cite{MOCO_2020_CVPR} &   a  & \xmark   & \xmark     & 0.35$\pm$0.02 & 0.24$\pm$0.01 & 0.00$\pm$0.00 & 0.44$\pm$0.01 & 0.49$\pm$0.04 \\    
& MOCO  \cite{MOCO_2020_CVPR} &   c  & \xmark    & \xmark     &  0.80$\pm$0.37 & 0.79$\pm$0.39 & 0.78$\pm$0.42 & 0.82$\pm$0.34 & 0.83$\pm$0.33\\    
& MOCO  \cite{MOCO_2020_CVPR} &   c  & \cmark    & \xmark     &  0.87$\pm$0.27 & 0.84$\pm$0.31 & 0.82$\pm$0.35 & 0.88$\pm$0.24 & 0.88$\pm$0.23 \\    
& MOCO  \cite{MOCO_2020_CVPR} &   c  & \xmark    & \cmark     &  {1.00$\pm$0.00} & {1.00$\pm$0.00} & {1.00$\pm$0.00} & {1.00$\pm$0.00} &{ 1.00$\pm$0.00} \\    
& MOCO  \cite{MOCO_2020_CVPR} &   c  & \cmark    & \cmark     &  0.98$\pm$0.07 & 0.96$\pm$0.11 & 0.95$\pm$0.15 & 0.98$\pm$0.06 & 0.98$\pm$0.05  \\ 
& ViT \cite{dosovitskiy2020vit} & c & \cmark    & \cmark & 0.98$\pm$0.03 & 0.97$\pm$0.05 & 0.96$\pm$0.07 & 0.98$\pm$0.03 & 0.98$\pm$0.03 \\   \hline
\end{tabular}
\vspace{1mm} \\
\begin{tabular}{p{0.3cm}p{2cm} p{0.5cm}p{0.5cm}p{0.5cm} p{1.4cm}p{1.4cm}p{1.4cm}p{1.4cm}p{1.4cm}p{1.4cm}p{1.4cm}p{1.4cm}|}\hline
\multirow{6}{*}{\rotatebox[origin=c]{90}{Plastic Mask}} 
& MOCO  \cite{MOCO_2020_CVPR} &   a  & \xmark    & \xmark     & 0.56$\pm$0.25 & 0.43$\pm$0.27 & 0.35$\pm$0.28 & 0.51$\pm$0.24 & 0.51$\pm$0.24\\  
& MOCO  \cite{MOCO_2020_CVPR} &   c  & \xmark    & \xmark     & 0.87$\pm$0.18 & 0.81$\pm$0.23 & 0.76$\pm$0.28 & 0.80$\pm$0.21 & 0.83$\pm$0.19\\    
& MOCO  \cite{MOCO_2020_CVPR} &   c  & \cmark    & \xmark     & 0.89$\pm$0.17 & 0.84$\pm$0.23 & 0.80$\pm$0.26 & 0.83$\pm$0.19 & 0.84$\pm$0.19  \\    
& MOCO  \cite{MOCO_2020_CVPR} &   c  & \xmark    & \cmark     & {0.92$\pm$0.15} & {0.88$\pm$0.21} & {0.83$\pm$0.25} & {0.85$\pm$0.18} & {0.87$\pm$0.17}\\    
& MOCO  \cite{MOCO_2020_CVPR} &   c  & \cmark    & \cmark     & 0.91$\pm$0.15 & 0.87$\pm$0.20 & 0.82$\pm$0.24 & 0.84$\pm$0.17 & 0.87$\pm$0.16\\  
& ViT \cite{dosovitskiy2020vit} & c & \cmark    & \cmark & 0.94$\pm$0.10 & 0.90$\pm$0.14 & 0.86$\pm$0.18 & 0.86$\pm$0.13 & 0.90$\pm$0.11\\  \hline
\end{tabular}
\vspace{1mm} \\
\begin{tabular}{p{0.3cm}p{2cm} p{0.5cm}p{0.5cm}p{0.5cm} p{1.4cm}p{1.4cm}p{1.4cm}p{1.4cm}p{1.4cm}p{1.4cm}p{1.4cm}p{1.4cm}|}\hline
\multirow{6}{*}{\rotatebox[origin=c]{90}{Makeup}} 
& MOCO  \cite{MOCO_2020_CVPR} &   a  & \xmark    & \xmark     &  0.42$\pm$0.40 & 0.36$\pm$0.42 & 0.30$\pm$0.45 & 0.41$\pm$0.36 & 0.38$\pm$0.37\\   
& MOCO  \cite{MOCO_2020_CVPR} &   c  & \xmark    & \xmark     &  0.44$\pm$0.41 & 0.35$\pm$0.45 & 0.33$\pm$0.46 & 0.41$\pm$0.39 & 0.39$\pm$0.40 \\    
& MOCO  \cite{MOCO_2020_CVPR} &   c  & \cmark    & \xmark     & 0.46$\pm$0.40 & 0.37$\pm$0.44 & 0.34$\pm$0.46 & 0.44$\pm$0.38 & 0.41$\pm$0.39\\   
& MOCO  \cite{MOCO_2020_CVPR} &   c  & \xmark    & \cmark     &  0.50$\pm$0.37 & 0.41$\pm$0.41 & 0.34$\pm$0.43 & 0.46$\pm$0.33 & 0.43$\pm$0.36 \\  
& MOCO  \cite{MOCO_2020_CVPR} &   c  & \cmark    & \cmark     &  {0.53$\pm$0.38} & {0.44$\pm$0.42} & 0{.37$\pm$0.45} & {0.49$\pm$0.35} & {0.47$\pm$0.37} \\
& ViT \cite{dosovitskiy2020vit} & c & \cmark    & \cmark & 0.60$\pm$0.35 & 0.52$\pm$0.39 & 0.44$\pm$0.44 & 0.56$\pm$0.32 & 0.54$\pm$0.35 \\   \hline
\end{tabular}
\vspace{1mm} \\
\begin{tabular}{p{0.3cm}p{2cm} p{0.5cm}p{0.5cm}p{0.5cm} p{1.4cm}p{1.4cm}p{1.4cm}p{1.4cm}p{1.4cm}p{1.4cm}p{1.4cm}p{1.4cm}|}\hline
\multirow{6}{*}{\rotatebox[origin=c]{90}{Silicone Mask}} 
& MOCO  \cite{MOCO_2020_CVPR} &   a  & \xmark    & \xmark     &   0.38$\pm$0.32 & 0.29$\pm$0.35 & 0.18$\pm$0.39 & 0.41$\pm$0.30 & 0.44$\pm$0.31 \\    
& MOCO  \cite{MOCO_2020_CVPR} &   c  & \xmark    & \xmark     &{ 1.00$\pm$0.00} & {1.00$\pm$0.00} & {1.00$\pm$0.00} & {1.00$\pm$0.00} & {1.00$\pm$0.00}\\    
& MOCO  \cite{MOCO_2020_CVPR} &   c  & \cmark    & \xmark     & {1.00$\pm$0.00} & {1.00$\pm$0.00} & {1.00$\pm$0.00} & {1.00$\pm$0.00} & {1.00$\pm$0.00}\\    
& MOCO  \cite{MOCO_2020_CVPR} &   c  & \xmark    & \cmark     & 0.91$\pm$0.28 & 0.91$\pm$0.29 & 0.91$\pm$0.29 & 0.92$\pm$0.26 & 0.92$\pm$0.26  \\    
& MOCO  \cite{MOCO_2020_CVPR} &   c  & \cmark    & \cmark     & 0.91$\pm$0.28 & 0.91$\pm$0.29 & 0.91$\pm$0.29 & 0.92$\pm$0.26 & 0.92$\pm$0.26 \\   
& ViT \cite{dosovitskiy2020vit} & c & \cmark    & \cmark & 1.00$\pm$0.00 & 1.00$\pm$0.00 & 1.00$\pm$0.00 & 1.00$\pm$0.00 & 1.00$\pm$0.00 \\   \hline
\end{tabular}
\vspace{1mm} \\
\begin{tabular}{p{0.3cm}p{2cm} p{0.5cm}p{0.5cm}p{0.5cm} p{1.4cm}p{1.4cm}p{1.4cm}p{1.4cm}p{1.4cm}p{1.4cm}p{1.4cm}p{1.4cm}|}\hline
\multirow{6}{*}{\rotatebox[origin=c]{90}{Paper Glasses}} 
& MOCO  \cite{MOCO_2020_CVPR} &   a  & \xmark    & \xmark     & 0.27$\pm$0.09 & 0.15$\pm$0.09 & 0.00$\pm$0.00 & 0.23$\pm$0.07 & 0.19$\pm$0.07   \\    
& MOCO  \cite{MOCO_2020_CVPR} &   c  & \xmark    & \xmark     & 0.94$\pm$0.18 & 0.91$\pm$0.21 & {0.89$\pm$0.22} & 0.94$\pm$0.17 & 0.93$\pm$0.18\\   
& MOCO  \cite{MOCO_2020_CVPR} &   c  & \cmark    & \xmark     & 0.92$\pm$0.17 & 0.88$\pm$0.21 & 0.85$\pm$0.22 & 0.92$\pm$0.16 & 0.91$\pm$0.18\\    
& MOCO  \cite{MOCO_2020_CVPR} &   c  & \xmark    & \cmark     &  {0.95$\pm$0.05} & 0.91$\pm$0.08 & 0.87$\pm$0.12 & {0.97$\pm$0.03} & 0.94$\pm$0.05\\   
& MOCO  \cite{MOCO_2020_CVPR} &   c  & \cmark    & \cmark     &  {0.95$\pm$0.05} & {0.92$\pm$0.08} & 0.88$\pm$0.12 & 0.96$\pm$0.04 & {0.95$\pm$0.05}\\  
& ViT \cite{dosovitskiy2020vit} & c & \cmark    & \cmark & 0.96$\pm$0.14 & 0.94$\pm$0.19 & 0.92$\pm$0.24 & 0.92$\pm$0.15 & 0.94$\pm$0.16 \\  \hline
\end{tabular}
\vspace{1mm} \\
\begin{tabular}{p{0.3cm}p{2cm} p{0.5cm}p{0.5cm}p{0.5cm} p{1.4cm}p{1.4cm}p{1.4cm}p{1.4cm}p{1.4cm}p{1.4cm}p{1.4cm}p{1.4cm}|}\hline
\multirow{6}{*}{\rotatebox[origin=c]{90}{Tattoo}} 
& MOCO  \cite{MOCO_2020_CVPR} &   a  & \xmark    & \xmark     &  0.16$\pm$0.16 & 0.04$\pm$0.08 & 0.00$\pm$0.00 & 0.19$\pm$0.10 & 0.16$\pm$0.10   \\    
& MOCO  \cite{MOCO_2020_CVPR} &   c  & \xmark    & \xmark     &  0.77$\pm$0.29 & 0.70$\pm$0.37 & 0.63$\pm$0.42 & 0.74$\pm$0.31 & 0.73$\pm$0.32\\    
& MOCO  \cite{MOCO_2020_CVPR} &   c  & \cmark    & \xmark     &  0.92$\pm$0.21 & 0.90$\pm$0.25 & 0.88$\pm$0.29 & 0.92$\pm$0.21 & 0.91$\pm$0.21 \\    
& MOCO  \cite{MOCO_2020_CVPR} &   c  & \xmark    & \cmark     &  {0.97$\pm$0.18} & {0.96$\pm$0.19} & {0.95$\pm$0.19} & 0.96$\pm$0.16 & 0.95$\pm$0.17 \\    
& MOCO  \cite{MOCO_2020_CVPR} &   c  & \cmark    & \cmark     &  {0.97$\pm$0.18 }& {0.96$\pm$0.19} & {0.95$\pm$0.19} & {0.97$\pm$0.16 }& {0.96$\pm$0.17} \\   
& ViT \cite{dosovitskiy2020vit} & c & \cmark    & \cmark & 0.99$\pm$0.04 & 0.98$\pm$0.07 & 0.97$\pm$0.10 & 0.98$\pm$0.10 & 0.97$\pm$0.11 \\  \hline
\end{tabular}
\end{table*}

Since our main objective in this work is to provide explanations over any existing face-PAD model without affecting PAD performance, we exclude graphical model \textit{B} in our experiments and use only graphical models \textit{A}  and \textit{C}, in which the parameters of the LG module $\Theta_e$ are learned while the parameters of the PAD module $\Theta_p$ are frozen.  LG numerical evaluations resulting from graphical models \textit{A} and \textit{C} are shown in 
the first and second rows of the sub-tables in \cref{tab:results} utilizing  word-wise loss  $\mathcal{L}_{ww}$ in the absence of the sentence discriminative $\mathcal{L}_{disc}$ and sentence semantic $\mathcal{L}_{ss}$ losses. In comparison to graphical model \textit{A}, it can be seen that graphical model \textit{C} consistently leads to better results whether considering all PA categories combined or separated.

The quantitative results within the third-to-fifth rows of the sub-tables correspond to different experiments in the presence or absence of the sentence-discriminative $\mathcal{L}_{disc}$ or sentence semantic  $\mathcal{L}_{ss}$ losses. The results verify usefulness of the presence of each of the losses and it can be observed that the best performance is achieved when both are included.

A comparison among the results of the different PA types indicates that makeup and paper glasses are the top two challenging PA categories for explainability. However, noting that the average of the predicted PAD scores for these two classes (0.92$\pm$0.21 and 0.93$\pm$0.13)  is lower than the corresponding statistics of other PA types, it becomes clear that these two PA types are challenging for PAD as well. Overall, it can be seen that the ViT model \cite{dosovitskiy2020vit} leads to superior performances on both PAD and LG.

Finally, in  \cref{fig:ex_desc}, we show qualitative examples of the generated descriptions for PA images. The first row shows examples of meaningful generated descriptions and the second row shows samples of incorrectly generated descriptions. It can be seen that the predicted PAD scores for the second row are not as high as the ones for the first row, which explains the poor descriptions in the second row since descriptions depend on PAD predictions in our model.


\section{Conclusion} 
\label{sec:conclusion}
We introduced a biometrics-NLP task that exploits the use of natural language for explaining face-PAD (XFace-PAD). To solve this challenging problem, we prepared a new face PAD dataset that is annotated with five descriptions per PA image. Because of having a limited number of annotations in our study, we leveraged pre-trained backbone models and only trained lightweight extensions. We used a lightweight LSTM network in our language generation model.
We investigated applying different language losses including word-wise cross-entropy loss, sentence-semantic loss based on BERT's sentence embedding, and  sentence discriminative losses.
Our results indicate the usefulness of including the sentence-wise losses. Since in the current work our main objective was to provide explanations through language for a given trained PAD model, the parameters of the PAD model were frozen while training the language generation model.
Investigating the possibility and effect of simultaneously training both the PAD and language generation models constitute part of our future work.

\noindent\textbf{Acknowledgement}
This research is based upon work supported by the Office of the Director of National Intelligence (ODNI), Intelligence Advanced Research Projects Activity (IARPA), via IARPA R\&D Contract No. 2017-17020200005. The views and conclusions contained herein should not be interpreted as necessarily representing the official policies or endorsements, either expressed or implied, of the ODNI, IARPA, or the U.S. Government.  The U.S. Government is authorized to reproduce and distribute reprints for Governmental purposes not withstanding any copyright annotation thereon.

{\small
\bibliographystyle{ieee}
\bibliography{egbib}

\begin{thebibliography}{10}\itemsep=-1pt

\bibitem{Akata_2013_CVPR}
Z.~Akata, F.~Perronnin, Z.~Harchaoui, and C.~Schmid.
\newblock Label-embedding for attribute-based classification.
\newblock In {\em 2013 IEEE Conference on Computer Vision and Pattern
  Recognition}, pages 819--826, 2013.

\bibitem{atoum2017face}
Y.~Atoum, Y.~Liu, A.~Jourabloo, and X.~Liu.
\newblock Face anti-spoofing using patch and depth-based cnns.
\newblock In {\em 2017 IEEE International Joint Conference on Biometrics
  (IJCB)}, pages 319--328. IEEE, 2017.

\bibitem{meteor_metric}
S.~Banerjee and A.~Lavie.
\newblock {METEOR}: An automatic metric for {MT} evaluation with improved
  correlation with human judgments.
\newblock In {\em Proceedings of the {ACL} Workshop on Intrinsic and Extrinsic
  Evaluation Measures for Machine Translation and/or Summarization}, pages
  65--72, Ann Arbor, Michigan, June 2005. Association for Computational
  Linguistics.

\bibitem{Rajarshi2020}
R.~Biswas, M.~Barz, and D.~Sonntag.
\newblock Towards explanatory interactive image captioning using top-down and
  bottom-up features, beam search and re-ranking.
\newblock {\em KI - Künstliche Intelligenz}, pages 1--14, 07 2020.

\bibitem{Boulkenafet2015}
Z.~{Boulkenafet}, J.~{Komulainen}, and A.~{Hadid}.
\newblock Face anti-spoofing based on color texture analysis.
\newblock In {\em 2015 IEEE International Conference on Image Processing
  (ICIP)}, pages 2636--2640, 2015.

\bibitem{SURF2016}
Z.~Boulkenafet, J.~Komulainen, and A.~Hadid.
\newblock Face anti-spoofing using speeded-up robust features and fisher vector
  encoding.
\newblock {\em IEEE Signal Processing Letters}, PP, 11 2016.

\bibitem{Chen_2021_WACV}
C.~Chen and A.~Ross.
\newblock An explainable attention-guided iris presentation attack detector.
\newblock In {\em Proceedings of the IEEE/CVF Winter Conference on Applications
  of Computer Vision (WACV) Workshops}, pages 97--106, January 2021.

\bibitem{chen2016scacnn}
L.~Chen, H.~Zhang, J.~Xiao, L.~Nie, J.~Shao, W.~Liu, and T.-S. Chua.
\newblock Sca-cnn: Spatial and channel-wise attention in convolutional networks
  for image captioning, 2016.

\bibitem{church_2017}
K.~W. CHURCH.
\newblock Word2vec.
\newblock {\em Natural Language Engineering}, 23(1):155–162, 2017.

\bibitem{Pereira2013}
T.~de~Freitas~Pereira, A.~Anjos, J.~M. De~Martino, and S.~Marcel.
\newblock Lbp-top based countermeasure against face spoofing attacks.
\newblock In J.-I. Park and J.~Kim, editors, {\em Computer Vision - ACCV 2012
  Workshops}, pages 121--132, Berlin, Heidelberg, 2013. Springer Berlin
  Heidelberg.

\bibitem{deng2009imagenet}
J.~Deng, W.~Dong, R.~Socher, L.-J. Li, K.~Li, and L.~Fei-Fei.
\newblock Imagenet: A large-scale hierarchical image database.
\newblock In {\em 2009 IEEE conference on computer vision and pattern
  recognition}, pages 248--255. Ieee, 2009.

\bibitem{devlin-etal-2019-bert}
J.~Devlin, M.-W. Chang, K.~Lee, and K.~Toutanova.
\newblock {BERT}: Pre-training of deep bidirectional transformers for language
  understanding.
\newblock In {\em Proceedings of the 2019 Conference of the North {A}merican
  Chapter of the Association for Computational Linguistics: Human Language
  Technologies, Volume 1 (Long and Short Papers)}, pages 4171--4186,
  Minneapolis, Minnesota, June 2019. Association for Computational Linguistics.

\bibitem{dosovitskiy2020vit}
A.~Dosovitskiy, L.~Beyer, A.~Kolesnikov, D.~Weissenborn, X.~Zhai,
  T.~Unterthiner, M.~Dehghani, M.~Minderer, G.~Heigold, S.~Gelly, J.~Uszkoreit,
  and N.~Houlsby.
\newblock An image is worth 16x16 words: Transformers for image recognition at
  scale.
\newblock {\em ICLR}, 2021.

\bibitem{bleu_metric}
M.~Dreyer and D.~Marcu.
\newblock {H}y{TER}: Meaning-equivalent semantics for translation evaluation.
\newblock In {\em Proceedings of the 2012 Conference of the North {A}merican
  Chapter of the Association for Computational Linguistics: Human Language
  Technologies}, pages 162--171, Montr{\'e}al, Canada, June 2012. Association
  for Computational Linguistics.

\bibitem{Fong2017}
R.~C. Fong and A.~Vedaldi.
\newblock Interpretable explanations of black boxes by meaningful perturbation.
\newblock {\em 2017 IEEE International Conference on Computer Vision (ICCV)},
  Oct 2017.

\bibitem{Gan2017}
J.~{Gan}, S.~{Li}, Y.~{Zhai}, and C.~{Liu}.
\newblock 3d convolutional neural network based on face anti-spoofing.
\newblock In {\em 2017 2nd International Conference on Multimedia and Image
  Processing (ICMIP)}, pages 1--5, 2017.

\bibitem{Anjith2021IJCB}
A.~George and S.~Marcel.
\newblock On the effectiveness of vision transformers for zero-shot face
  anti-spoofing.
\newblock {\em IJCB}, abs/2011.08019, 2021.

\bibitem{LBP2015}
D.~{Gragnaniello}, G.~{Poggi}, C.~{Sansone}, and L.~{Verdoliva}.
\newblock An investigation of local descriptors for biometric spoofing
  detection.
\newblock {\em IEEE Transactions on Information Forensics and Security},
  10(4):849--863, 2015.

\bibitem{MOCO_2020_CVPR}
K.~He, H.~Fan, Y.~Wu, S.~Xie, and R.~Girshick.
\newblock Momentum contrast for unsupervised visual representation learning.
\newblock In {\em Proceedings of the IEEE/CVF Conference on Computer Vision and
  Pattern Recognition (CVPR)}, June 2020.

\bibitem{he2016resnet}
K.~He, X.~Zhang, S.~Ren, and J.~Sun.
\newblock Deep residual learning for image recognition.
\newblock In {\em 2016 IEEE Conference on Computer Vision and Pattern
  Recognition (CVPR)}, pages 770--778, 2016.

\bibitem{he2020image}
S.~He, W.~Liao, H.~R. Tavakoli, M.~Yang, B.~Rosenhahn, and N.~Pugeault.
\newblock Image captioning through image transformer, 2020.

\bibitem{hendricks2016generating}
L.~A. Hendricks, Z.~Akata, M.~Rohrbach, J.~Donahue, B.~Schiele, and T.~Darrell.
\newblock Generating visual explanations.
\newblock {\em Proceedings of the European Conference on Computer Vision
  (ECCV)}, 2016.

\bibitem{hu2017squeezeandexcitation}
J.~Hu, L.~Shen, S.~Albanie, G.~Sun, and E.~Wu.
\newblock Squeeze-and-excitation networks, 2017.

\bibitem{jaiswal2019ropad}
A.~Jaiswal, S.~Xia, I.~Masi, and W.~AbdAlmageed.
\newblock Ropad: Robust presentation attack detection through unsupervised
  adversarial invariance, 2019.

\bibitem{jourabloo2018face}
A.~Jourabloo, Y.~Liu, and X.~Liu.
\newblock Face de-spoofing: Anti-spoofing via noise modeling.
\newblock In {\em Proceedings of the European Conference on Computer Vision
  (ECCV)}, pages 290--306, 2018.

\bibitem{Adam}
D.~P. Kingma and J.~Ba.
\newblock Adam: A method for stochastic optimization, 2014.
\newblock cite arxiv:1412.6980Comment: Published as a conference paper at the
  3rd International Conference for Learning Representations, San Diego, 2015.

\bibitem{Komulainen2013}
J.~{Komulainen}, A.~{Hadid}, and M.~{Pietikäinen}.
\newblock Context based face anti-spoofing.
\newblock In {\em 2013 IEEE Sixth International Conference on Biometrics:
  Theory, Applications and Systems (BTAS)}, pages 1--8, 2013.

\bibitem{Lampert_2009_CVPR}
C.~H. Lampert, H.~Nickisch, and S.~Harmeling.
\newblock Learning to detect unseen object classes by between-class attribute
  transfer.
\newblock In {\em 2009 IEEE Conference on Computer Vision and Pattern
  Recognition}, pages 951--958, 2009.

\bibitem{lin-2004-rouge}
C.-Y. Lin.
\newblock {ROUGE}: A package for automatic evaluation of summaries.
\newblock In {\em Text Summarization Branches Out}, pages 74--81, Barcelona,
  Spain, July 2004. Association for Computational Linguistics.

\bibitem{Liu2018}
Y.~{Liu}, A.~{Jourabloo}, and X.~{Liu}.
\newblock Learning deep models for face anti-spoofing: Binary or auxiliary
  supervision.
\newblock In {\em 2018 IEEE/CVF Conference on Computer Vision and Pattern
  Recognition}, pages 389--398, 2018.

\bibitem{liu2018learning}
Y.~Liu, A.~Jourabloo, and X.~Liu.
\newblock Learning deep models for face anti-spoofing: Binary or auxiliary
  supervision.
\newblock In {\em Proceedings of the IEEE Conference on Computer Vision and
  Pattern Recognition}, pages 389--398, 2018.

\bibitem{Miller1990}
G.~A. Miller, R.~Beckwith, C.~Fellbaum, D.~Gross, and K.~J. Miller.
\newblock {Introduction to WordNet: an on-line lexical database}.
\newblock {\em International Journal of Lexicography}, 3(4):235--244, 1990.

\bibitem{Patel2016}
K.~{Patel}, H.~{Han}, and A.~K. {Jain}.
\newblock Secure face unlock: Spoof detection on smartphones.
\newblock {\em IEEE Transactions on Information Forensics and Security},
  11(10):2268--2283, 2016.

\bibitem{perez2019deep}
D.~P{\'e}rez-Cabo, D.~Jim{\'e}nez-Cabello, A.~Costa-Pazo, and R.~J.
  L{\'o}pez-Sastre.
\newblock Deep anomaly detection for generalized face anti-spoofing.
\newblock In {\em Proceedings of the IEEE Conference on Computer Vision and
  Pattern Recognition Workshops}, pages 0--0, 2019.

\bibitem{Qin2020}
Y.~Qin, C.~Zhao, X.~Zhu, Z.~Wang, Z.~Yu, T.~Fu, F.~Zhou, J.~Shi, and Z.~Lei.
\newblock Learning meta model for zero- and few-shot face anti-spoofing.
\newblock {\em Proceedings of the AAAI Conference on Artificial Intelligence},
  34:11916--11923, 04 2020.

\bibitem{ribeiro2016why}
M.~T. Ribeiro, S.~Singh, and C.~Guestrin.
\newblock "why should i trust you?": Explaining the predictions of any
  classifier, 2016.

\bibitem{Rostami_2021_ICCV}
M.~Rostami, L.~Spinoulas, M.~Hussein, J.~Mathai, and W.~Abd-Almageed.
\newblock Detection and continual learning of novel face presentation attacks.
\newblock In {\em Proceedings of the IEEE/CVF International Conference on
  Computer Vision (ICCV)}, pages 14851--14860, October 2021.

\bibitem{russakovskyImageNetLargeScale2015}
O.~Russakovsky, J.~Deng, H.~Su, J.~Krause, S.~Satheesh, S.~Ma, Z.~Huang,
  A.~Karpathy, A.~Khosla, M.~Bernstein, A.~C. Berg, and L.~Fei-Fei.
\newblock {ImageNet} {Large} {Scale} {Visual} {Recognition} {Challenge}.
\newblock {\em International Journal of Computer Vision}, 115(3):211--252, Dec.
  2015.

\bibitem{selvaraju2016gradcam}
R.~R. Selvaraju, A.~Das, R.~Vedantam, M.~Cogswell, D.~Parikh, and D.~Batra.
\newblock Grad-cam: Why did you say that?, 2016.

\bibitem{Silva_RECPAD_2020}
W.~Silva, J.~Ribeiro~Pinto, T.~Gonçalves, A.~Sequeira, and J.~Cardoso.
\newblock Explainable artificial intelligence for face presentation attack
  detection.
\newblock In {\em 26th Portuguese Conference in Pattern Recognition (RECPAD)},
  10 2020.

\bibitem{simonyan2014deep}
K.~Simonyan, A.~Vedaldi, and A.~Zisserman.
\newblock Deep inside convolutional networks: Visualising image classification
  models and saliency maps, 2014.

\bibitem{spinoulas2020multispectral}
L.~Spinoulas, M.~Hussein, D.~Geissbühler, J.~Mathai, O.~G. Almeida, G.~Clivaz,
  S.~Marcel, and W.~AbdAlmageed.
\newblock Multispectral biometrics system framework: Application to
  presentation attack detection, 2020.

\bibitem{Tsai_2017_ICCV}
Y.-H.~H. Tsai, L.-K. Huang, and R.~Salakhutdinov.
\newblock Learning robust visual-semantic embeddings.
\newblock In {\em 2017 IEEE International Conference on Computer Vision
  (ICCV)}, pages 3591--3600, 2017.

\bibitem{vilone2020explainable}
G.~Vilone and L.~Longo.
\newblock Explainable artificial intelligence: a systematic review, 2020.

\bibitem{wang2020deep}
Z.~Wang, Z.~Yu, C.~Zhao, X.~Zhu, Y.~Qin, Q.~Zhou, F.~Zhou, and Z.~Lei.
\newblock Deep spatial gradient and temporal depth learning for face
  anti-spoofing.
\newblock In {\em Proceedings of the IEEE/CVF Conference on Computer Vision and
  Pattern Recognition}, pages 5042--5051, 2020.

\bibitem{williford2020explainable}
J.~R. Williford, B.~B. May, and J.~Byrne.
\newblock Explainable face recognition, 2020.

\bibitem{pmlr-v37-xuc15}
K.~Xu, J.~Ba, R.~Kiros, K.~Cho, A.~Courville, R.~Salakhudinov, R.~Zemel, and
  Y.~Bengio.
\newblock Show, attend and tell: Neural image caption generation with visual
  attention.
\newblock In F.~Bach and D.~Blei, editors, {\em Proceedings of the 32nd
  International Conference on Machine Learning}, volume~37 of {\em Proceedings
  of Machine Learning Research}, pages 2048--2057, Lille, France, 07--09 Jul
  2015. PMLR.

\bibitem{yang2014learn}
J.~Yang, Z.~Lei, and S.~Z. Li.
\newblock Learn convolutional neural network for face anti-spoofing, 2014.

\bibitem{yang2019face}
X.~Yang, W.~Luo, L.~Bao, Y.~Gao, D.~Gong, S.~Zheng, Z.~Li, and W.~Liu.
\newblock Face anti-spoofing: Model matters, so does data.
\newblock In {\em Proceedings of the IEEE Conference on Computer Vision and
  Pattern Recognition}, pages 3507--3516, 2019.

\bibitem{Yu2020}
Z.~{Yu}, Y.~{Qin}, X.~{Xu}, C.~{Zhao}, Z.~{Wang}, Z.~{Lei}, and G.~{Zhao}.
\newblock Auto-fas: Searching lightweight networks for face anti-spoofing.
\newblock In {\em ICASSP 2020 - 2020 IEEE International Conference on
  Acoustics, Speech and Signal Processing (ICASSP)}, pages 996--1000, 2020.

\bibitem{yu2020searching}
Z.~Yu, C.~Zhao, Z.~Wang, Y.~Qin, Z.~Su, X.~Li, F.~Zhou, and G.~Zhao.
\newblock Searching central difference convolutional networks for face
  anti-spoofing.
\newblock In {\em Proceedings of the IEEE/CVF Conference on Computer Vision and
  Pattern Recognition}, pages 5295--5305, 2020.

\bibitem{zeiler2013visualizing}
M.~D. Zeiler and R.~Fergus.
\newblock Visualizing and understanding convolutional networks, 2013.

\bibitem{Zhang_2017_CVPR}
L.~Zhang, T.~Xiang, and S.~Gong.
\newblock Learning a deep embedding model for zero-shot learning.
\newblock In {\em Proceedings of the IEEE Conference on Computer Vision and
  Pattern Recognition (CVPR)}, July 2017.

\bibitem{zhang2018image}
Y.~Zhang, K.~Li, K.~Li, L.~Wang, B.~Zhong, and Y.~Fu.
\newblock Image super-resolution using very deep residual channel attention
  networks, 2018.

\bibitem{zhou2015learning}
B.~Zhou, A.~Khosla, A.~Lapedriza, A.~Oliva, and A.~Torralba.
\newblock Learning deep features for discriminative localization, 2015.

\end{thebibliography}
}

\end{document}